\documentclass[11pt]{article}
\usepackage[utf8]{inputenc}
\usepackage{geometry}
\usepackage{amsmath}
\usepackage{dsfont}
\usepackage{setspace}
\usepackage[breakable]{tcolorbox}
\usepackage{dirtytalk}
\usepackage{algorithm}
\usepackage{algpseudocode}
\usepackage{graphicx}
\usepackage{capt-of}
\usepackage[symbol]{footmisc}

\DeclareMathOperator*{\argmin}{arg\min}

\DeclareMathOperator{\sign}{sgn}
\usepackage{authblk}
\usepackage{hyperref}
\hypersetup{
    colorlinks=true,
    linkcolor=black,
    filecolor=magenta,      
    urlcolor=cyan,
}

\title{\fontseries{b}\selectfont Deep Unrolled Recovery in Sparse Biological Imaging}
\author[1]{Yair Ben Sahel\thanks{These authors contributed equally}}
\author[2]{John P. Bryan\textsuperscript{\textasteriskcentered}}
\author[2]{Brian Cleary}
\author[2]{Samouil L. Farhi}
\author[1]{Yonina C. Eldar}
\affil[1]{\small Department of Computer Science and Applied Mathematics, Weizmann Institute of Science, Israel}
\affil[2]{\small Broad Institute of MIT and Harvard}
\date{}

\begin{document}

\maketitle

\vspace{-2.5em}

\begin{center}\textit{Deep algorithm unrolling has emerged as a powerful model-based approach to develop deep architectures that combine the interpretability of iterative algorithms with the performance gains of supervised deep learning, especially in cases of sparse optimization. This framework is well-suited to applications in biological imaging, where physics-based models exist to describe the measurement process and the information to be recovered is often highly structured. Here, we review the method of deep unrolling, and show how it improves source localization in several biological imaging settings.}\end{center}


\section{Introduction}
Biological imaging which precisely labels microscopic structures offers a range of insights, from Single Molecule Localization Microscopy (SMLM) \cite{storm_09} visualizing the microarchitecture of the cellular cytoskeleton; to single-molecule Fluorescence \emph{in situ} Hybridization \cite{smfish} revealing the spatial distribution of gene expression; and synaptic immunofluorescence \cite{dogNet} showing the distribution of neuronal connections in the brain. To succeed, each of these imaging methods is accompanied by analysis tools to identify and localize the desired signal. 

\begin{figure}
    \centering
    \includegraphics[width=.5\textwidth,height=.2\textheight,keepaspectratio]{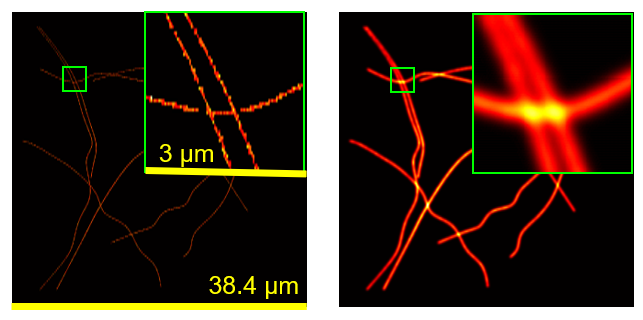}
    \caption{Simulated microscopic imaging of microtubules from \cite{LSPARCOM}. LEFT: True localization of microtubule structure. RIGHT: Imaged microtubule, ``blurred'' by the microscope PSF.}
    \label{fig:smlm}
\end{figure}

In localization problems, regardless of the imaging system used, analysis pipelines attempt to undo the blurring effect of its imperfect impulse response, or point-spread function (PSF). The PSF always has finite width, leading to limited image resolution. If small objects, like the two thin microtubules seen in Fig. \ref{fig:smlm},  are nearer to each other than the width of the PSF, they may be difficult to distinguish. The biological signal in the field of view (FOV) of the imaging system can be represented as a high resolution matrix, where each element's value represents the intensity of the signal at that physical location.  The imaging process may be thought of as a 2D convolution of the \say{true} object being imaged and an array representing the PSF. The goal of a localization algorithm is to \say{undo} this convolution. 

The localization problem is much more tractable if the images have predictable structure, because the solution space can be constrained. For example, some biological images are comprised of similarly-sized cells, elongated fibers of known width, or small, scattered fluorescent spots representing individual molecules with dimensions below the diffraction limit. This knowledge can be combined with an understanding of the physical parameters of the imaging system, such as the numerical aperture and magnification, in analysis pipelines, to identify cell centers \cite{ecnncs}, trace long fibers  \cite{storm_09}, or localize fluorophores which have been bound to biologically relevant molecules \cite{MERFISH}. The results of these algorithms may then be used to achieve higher level biological goals, from generating tissue atlases, to determining genetic expression patterns or diagnosing pathologies.

Here, we focus specifically on biological localization problems on sparse images, meaning that the high-resolution information to be recovered has relatively few nonzero values. Some signals, such as fluorescently tagged messenger RNA (mRNA), clusters of proteins, or micro-bubbles in ultrasound imaging are naturally sparse. In other cases, experimental techniques may be used to induce sparsity \cite{storm_09}, or sparsity in bases other than the spatial domain may be exploited.

Iterative optimization techniques have emerged as one of the most powerful approaches for localizing sparse signal emitters. For example, in SMLM, a super-resolution technique which relies on sub-pixel localization of scattered fluorophores, the original peak-finding algorithms have been outperformed by approaches using iterative convex optimization-based algorithms in terms of  localization accuracy, signal-to-noise ratio (SNR), and resolution  \cite{smlm_rev}. These advantages of iterative optimization techniques for sparse recovery go beyond SMLM to other biological imaging problems, providing high-accuracy localization in many settings \cite{JSIT}\cite{ULM}\cite{CISTA}. However, they also suffer some disadvantages. They require adjustment of optimization parameters and explicit knowledge of the impulse response of the imaging system, which restricts their use when the imaging system is not well-characterized. They are computationally expensive and converge slowly, which limits their use in real-time, live-cell imaging. Finally, they are relatively inflexible: a given algorithm is designed to take advantage of a particular structure (here, signal sparsity), but ignores other context which may be important (e.g., cell size or density).

Many of these disadvantages can be overcome by replacing the iterations of these algorithms with trained neural networks which perform the same mathematical operation, a process known as algorithmic unrolling \cite{LISTA} (alternatively, \say{unfolding}). By doing so, parameters which would have to be specified explicitly or tuned empirically are learned automatically, and relevant context ignored by the algorithm may be incorporated into the learned model. Since its introduction a decade ago, a wide variety of techniques have been adapted using learned unrolling, enabling improvements in performance across a variety of settings \cite{Unrolling_rev}.

In the rest of this paper, we review how learned unrolling is applied to the localization of sparse sources in biological imaging data. We first formulate biological localization as a sparse recovery problem, and discuss the advantages and disadvantages of iterative approaches to sparse recovery. We then describe how algorithmic unrolling addresses some of the shortcomings, and how the general sparse recovery problem may be adapted to the unrolling framework. Next we show in detail how unrolling has been used to achieve fast, accurate super-resolution in the optical microscopy technique SMLM. We then turn to review a number of additional biological imaging analysis problems to which unrolling has been applied to improve performance: Ultrasound Localization Microscopy (ULM), Light-Field Microscopy (LFM), and cell center localization in fluorescence microscopy. Throughout, we discuss a number of additional data analysis problems in sparse optical microscopy, and propose that algorithmic unrolling be applied, to achieve the same benefits obtained in the reviewed techniques.


\section{Sparse Recovery in Biological Imaging}
The localization of biological objects, from microtubules to neural synapses, can be approached effectively as a convex optimization problem. For ease of notation, we first reframe the imaging process, typically thought of as a convolution, as a matrix-vector multiplication. The high-resolution signal is \say{vectorized} and then multiplied by a matrix representing the PSF. We note that the problem can be formulated and solved with 2D convolutions equally well, and the techniques described throughout the paper may be applied to a 2D formulation, as in \cite{LSPARCOM} and \cite{CISTA}.

Formally, we consider the FOV as a high-resolution square grid with side length $n_{h}$. The total number of locations in this grid is $N_{h} = n_{h}^{2}$. This grid is \say{vectorized} to form a vector $\mathbf{x}\in\mathds{R}^{N_{h}}$. The locations of emitters in the sample may be modeled by assigning each element of $\mathbf{x}$ a value related to the number of photons emitted from that location within the FOV. If the FOV is imaged using a sensor with $N_{l}$ pixels (with $N_{h}>N_{l}$), then we can model the imaging process as multiplication by a matrix $\mathbf{A}\in\mathds{R}^{N_{l}\times N_{h}}$, in which element $(i,j)$ is the proportion of signal emitted from location $j$ on the high-resolution grid that will be detected at pixel $i$ of the sensor. Thus defined, the columns of $\mathbf{A}$ represent the PSF of the imaging system, such that column $j$ of $\mathbf{A}$ is the PSF of the system for a point source at location $j$. The (vectorized) measured image is then $\mathbf{y} = \mathbf{Ax}$, with $\mathbf{y}\in\mathds{R}^{N_{l}}$. The goal of the analysis pipeline is to infer the value of $\mathbf{x}$, given $\mathbf{y}$ and $\mathbf{A}$.

This inference problem can be formulated as a least-squares optimization problem: we seek to find 
\begin{equation}
\label{eqn:ls}
\mathbf{\hat{x}} = \argmin_{\mathbf{x}}\|\mathbf{y}-\mathbf{Ax}\|^{2}_{2}.
\end{equation}
Even if $\mathbf{A}$ is known perfectly, as long as $N_{h}>N_{l}$, $\mathbf{A}$ will have a nontrivial null space, so that the optimization problem is underdetermined. Leveraging knowledge of the biological structure of $\mathbf{x}$ can resolve this issue. If, as discussed above, it is known that $\mathbf{x}$ is sparse, then we may choose a sparse optimization technique, such as the well-known LASSO \cite{LASSO}, to recover $\mathbf{x}$:
\begin{equation}
\label{eqn:sls}
\mathbf{\hat{x}} = \argmin_{\mathbf{x}}\|\mathbf{y}-\mathbf{Ax}\|^{2}_{2}+\lambda\|\mathbf{x}\|_{1}.
\end{equation}
In particular, by correctly tuning $\lambda$, the minimizer $\mathbf{\hat{x}}$ of (\ref{eqn:sls}) will provide accurate locations of each signal-emitting object in the FOV.


\section{Algorithmic Unrolling for Sparse Localization}

Once a problem is framed as a sparse optimization of the form (\ref{eqn:sls}), a number of algorithms may be used to find the minimizer $\mathbf{\hat{x}}$. Examples include the Alternating Direction Method of Multipliers (ADMM), \cite{ADMM} the Iterative Shrinkage-Thresholding Algorithm (ISTA) \cite{ISTA}, and the Half-Quadratic Splitting algorithm (HQS)\cite{HQS-net}. These methods converge to the correct minimizer $\mathbf{\hat{x}}$, but have some limiting disadvantages, as discussed above: slow convergence, the requirement of parameters tuning and explicit knowledge of the imaging system \cite{LSPARCOM}, and mathematical inflexibility. 

Deep learning approaches have overcome some of these disadvantages. In analysis of SMLM data, convolutional neural network models have achieved fast, accurate super-resolution \cite{deepSTORM}, able to improve recovery by incorporating structures not specified by the user. Deep learning, however, comes with disadvantages of its own. In particular, deep learning is typically thought of as a \say{black box} process: it is difficult to interpret the way the model transforms input to obtain a result. Because of this, when inaccurate results are produced, it can be difficult to understand how to improve the model. Typical deep learning approaches are strongly dependent on the available training data, causing a lack of model robustness to new examples. Finally, when using generic network architectures, many layers and parameters are typically required for good performance.

In 2010, Gregor and LeCun proposed a method to create neural networks based on iterative methods used for sparse recovery \cite{LISTA}, known as algorithm unrolling. The goal is to take advantage of both the interpretability of iterative techniques and the flexibility of learned methods. In learned unrolling, the transformation applied to the input by each iteration of the algorithm is replaced with a neural network layer which applies the same type of function: for instance, matrix multiplication can be replaced by a fully-connected layer, and thresholding can be replaced by an activation function representing an appropriate regularizer. These iteration-layers are concatenated together, and the resulting model-based neural network is optimized using supervised learning, with training data consisting of paired examples of the signal vector $\mathbf{x}$ and measurement vector $\mathbf{y}$ from (\ref{eqn:sls}). Training data may be obtained, for example, from measurement simulations with known ground truth as in \cite{LSPARCOM}. A forward pass through the optimized network will then perform the same operations as the iterative algorithm, with the parameters of each transformation optimized to map the training input $\mathbf{y}$ to its paired signal $\mathbf{x}$.

Gregor and LeCun applied the unrolling framework to ISTA, calling the ISTA-inspired network \say{Learned ISTA}, or LISTA. For a given number of iterations/layers, the trained LISTA network obtains lower prediction error than ISTA, and even achieves faster convergence and higher accuracy than the accelerated version of ISTA, FISTA  \cite{LISTA}. In the \say{From ISTA to LISTA} box below, we detail the process of constructing the LISTA network based on ISTA.

\vspace{0.5em}

\begin{tcolorbox}[breakable,title={From ISTA to LISTA}]
\setstretch{1}
Here, we detail ISTA and use it as a case study to describe the process of algorithm unrolling. Given a problem with the form of (\ref{eqn:sls}), ISTA estimates $\mathbf{x}$, taking as inputs the measurement matrix $\mathbf{A}$, the measurement vector $\mathbf{y}$, the regularization parameter $\lambda$, and $L$, a Lipschitz constant of $\nabla \|\mathbf{Ax-y}\|_{2}^{2}$.
\begin{algorithm}[H]
\setstretch{1}
\caption{ISTA}
\label{alg:ISTA}
\begin{algorithmic}[1]
\Require{$\mathbf{y}$, $\mathbf{A}$, $\lambda$, $L$, number of iterations $k_{max}$ }
\Ensure{$\mathbf{\hat{x}}$}
\State $\mathbf{\hat{x}_{1}} = 0$, $k = 1$.
\While{$k<k_{max}$}
\State $\mathbf{\hat{x}_{k+1}} = \mathbf{\mathcal{T}}_{\frac{\lambda}{L}}(\mathbf{\hat{x}_{k}}-2L\mathbf{A^{T}}(\mathbf{A\hat{x}_{k}-y}))$
\State $k\leftarrow k+1$
\EndWhile
\State $\mathbf{\hat{x}} = \mathbf{\hat{x}_{k_{max}}}$

\end{algorithmic}
\end{algorithm}

Here, $\mathbf{\mathcal{T}}_{\frac{\lambda}{L}}$ represents the soft thresholding operator after which ISTA is named,
\begin{equation}
\label{eqn:th_op}
    \mathcal{T}_{\alpha}(x) = \max\{|x|-\alpha,0\}\cdot\sign(x),
\end{equation}
where $\sign(\cdot)$ is the sign operator,
\begin{equation}
    \sign(x) = 
    \begin{cases}
        -1, & x<0\\
        1, & x>0.
    \end{cases}
\end{equation}

The iterative step of ISTA is given in line 3 of Algorithm \ref{alg:ISTA}. The argument of $\mathbf{\mathcal{T}}_{\frac{\lambda}{L}}(\cdot)$ in the iterative step can be rewritten as the sum of matrix-vector products with $\mathbf{y}$ and $\mathbf{x}_{k}$:
\begin{equation}
    \mathbf{x_{k}}-2L\mathbf{A}^{T}(\mathbf{Ax_{k}-y})) = 2L\mathbf{A}^{T}\mathbf{y} + (\mathbf{I}-2L\mathbf{A}^{T}\mathbf{A})\mathbf{x_{k}} = \mathbf{W_{0k}y} + \mathbf{W_{k}x_{k}}.
\label{eqn:ISTA_step}
\end{equation}
This step can be modeled by the sum of fully-connected neural network layers and an activation function with learned threshold, as depicted in Fig. \ref{fig:lista}. By stringing several of these layers together, the resulting deep neural network, LISTA, has the same form as the operation performed by running ISTA over multiple iterations. Differently from ISTA, the weights of each layer are trained independently, providing greater flexibility.

\vspace{1em}

\centering
\includegraphics[scale=0.65]{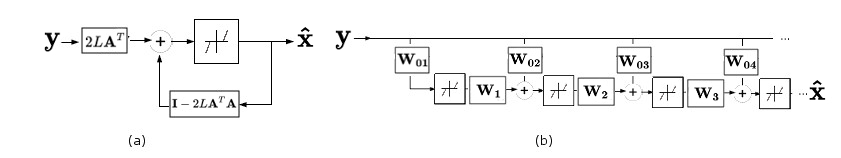}
\captionof{figure}{\footnotesize Unrolling of ISTA into LISTA. \textbf{(a)}: Diagram of the operation of ISTA as a feedback loop. \textbf{(b)}: Diagram of LISTA, matrix multiplications by $2L\mathbf{A}^{T}$ are replaced by the weight matrices, $\mathbf{W_{0k}}$, and multiplication by $\mathbf{I}-2L\mathbf{A}^{T}\mathbf{A}$ is replaced by $\mathbf{W_{k}}$, which, along with $\mathbf{W_{0k}}$, may be optimized with supervised learning.}
\label{fig:lista}

\end{tcolorbox}

This framework provides several key advantages: first, algorithm parameters, such as $\lambda$ in (\ref{eqn:sls}), are learned automatically. Second, with an unrolled model, the part of the model corresponding to $\mathbf{A}$ in (\ref{eqn:sls}), is learned, removing the need to explicitly model the PSF. Finally, while these iterative algorithms are designed to solve a specific problem, sparse recovery, the approach is general, solving all problems of this type equally well. With a neural network, the model can learn to analyze data that may have additional structure not explainable by sparsity, and thereby obtain higher-accuracy results more quickly. Because the underlying structure of the algorithm remains intact, the network is also less prone to overfitting, improving robustness. 

The unrolling framework also has a few drawbacks. The data-driven approach requires a substantial quantity of training data, which may be difficult to obtain. If the data used to train the network is generated differently from that being analyzed (for example, if a different microscope is used, with a substantially different PSF), recovery performance will degrade. However, it has been found that learned unrolled networks are much more robust than traditional learned neural networks to changes in the distribution of signal (for example, studying a different type of subcellular structure \cite{LSPARCOM}). The learned weights of the network may also be less interpretable than an algorithm's iterative step, in which each component has explicit physical meaning. In \cite{LSPARCOM}, however, it is shown that the LISTA-based neural network used for the superresolution task learns transformations which are closely related to the operation of the iterative step (for instance, convolution filters are learned with shapes similar to the PSF) and are easier to understand than those learned by the purely data-driven approach of typical neural networks. It is also important to note that, being partially data-driven, learned unrolled networks no longer explicitly solve the optimization problems they are based on (such as sparse recovery). While the structure of the algorithm is maintained, the transformation applied by the network does not follow the exact steps that are guaranteed to find the minimizer of (\ref{eqn:sls}), and may not be applied in the image domain. So, while learned unrolled networks have been shown to be successful in localizing spatially sparse sources, they do not explicitly solve the sparse recovery problem.

Throughout the rest of the paper, we will concentrate on applications of deep unrolling to the recovery of sparse biological data, specifically using unrolled networks based on ISTA. Importantly, the learned unrolling strategy is not restricted to ISTA, nor is it restricted to problems with a sparse prior: any algorithm for which the iterative step may be carried out by a learnable neural network layer may be unrolled. Gregor and LeCun developed a learned version of the coordinate descent algorithm, finding that the learned version again obtained much lower prediction error than the iterative version\cite{LISTA}. Other authors have applied the unrolling framework to a variety of algorithms for biological data processing tasks, including ADMM \cite{ADMM-net} and robust PCA \cite{rPCA}, which were shown to obtain lower errors in MRI signal recovery and ultrasound clutter suppression, respectively, in less time than the then state-of-the-art algorithms, consistent with learned unrolled networks converging more quickly. Outside of the realm of biology, in natural images unrolling of the half-quadratic splitting algorithm has been shown to achieve both high-quality denoising \cite{iminpaint} and super-resolution in natural images\cite{HQS-net}. Many additional example applications are provided in a recent review \cite{Unrolling_rev}.


\section{Unrolling in Optical Localization Microscopy}

In this section, we will focus on the domain of optical localization microscopy. First, we will give a detailed example of how unrolling enhances the capabilities of one optical imaging technique: Single-Molecule Localization Microscopy. Then, we will discuss how the concept of unrolling can be applied to other sparse biological optical imaging problems. 

\subsection{Unrolling in Single-Molecule Localization Microscopy}

Visualization of sub-cellular features and organelles within biological cells requires imaging techniques with nanometer resolution. In the case of optical imaging systems, from the 19\textsuperscript{th} century until the  recent development of super-resolution microscopy, the resolution limit was considered to be set by Abbe's diffraction limit for a microscope:

\begin{equation}
\label{eqn:difflim} 
    d=\frac{\beta}{2NA},
\end{equation}
where $d$ is the minimal distance below which two point sources cannot be distinguished, $\beta$ is the wavelength of the emitted photons, and NA is the numerical aperture of the microscope. In fluorescence microscopy, the sample is stained with fluorophores which can be excited with one color of light and emit photons of a higher wavelength for subsequent detection. Since most cells are not naturally fluorescent, this allows specific imaging of the stained biomolecules. If the number of photons emitted is sufficiently high, and the background is sufficiently low, single molecules can be detected in this way. However, biological structures of interest are typically made of multitudes of the same biomolecule type in close apposition, obscuring details finer than the diffraction limit of the emitted photons when all fluorophores are emitting at the same time.

One may overcome the diffraction limit by distinguishing between the photons coming from two neighboring fluorophores \cite{sub_diff}. One way to distinguish neighboring molecules, is by utilizing photo-activated or photo-switching fluorophores to separate fluorescent emission in time; this is the basis for SMLM techniques such as Photo-Activated Localization Microscopy (PALM) and Stochastic Optical Reconstruction Microscopy (STORM) \cite{SMLM1, SMLM2}. Optical, physical, or chemical means are used to ensure that at any given moment only a small subset of all flurophores are emitting photons. Then a large number of diffraction-limited images is collected, each containing just a few active isolated fluorophores. The imaging sequence is long enough such that each fluorophore is stochastically activated from a non-emissive state to a bright state, and back to a non-emissive (or bleached) state. During each cycle, the density of activated molecules is kept low enough that emission profiles of individual fluorophores do not overlap. 

\begin{figure}
    \centering
    \includegraphics[width=.7\textwidth,height=.3\textheight,keepaspectratio]{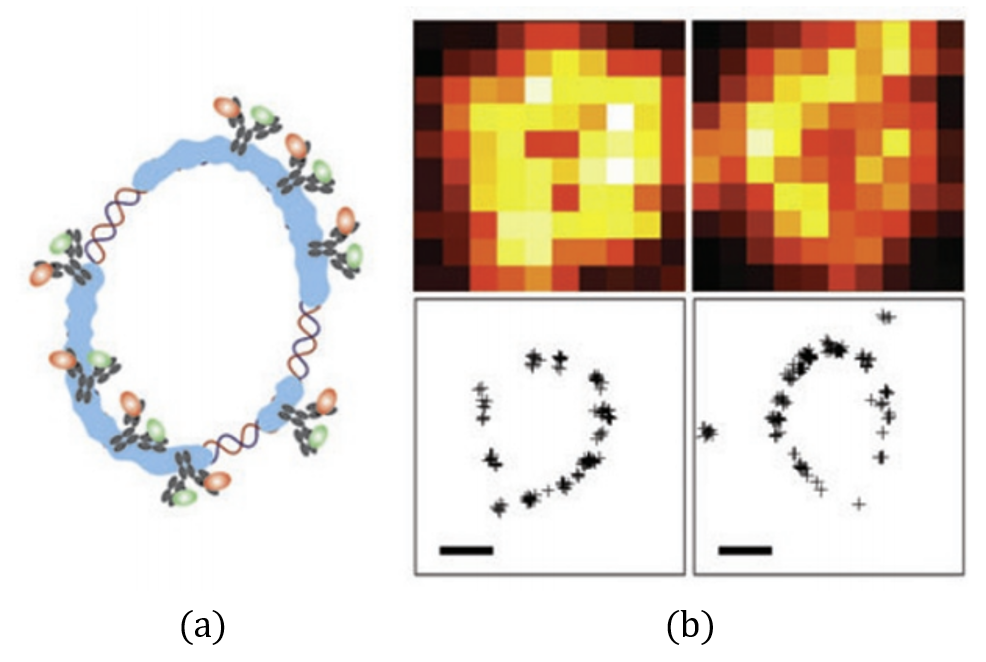}
    \caption{\footnotesize Sample experimental results from \cite{SMLM2}, comparing diffraction-limited and STORM-generated images of RecA-coated circular plasmid DNA. (a) Illustration of the DNA construct, with linked fluorophores (via immunohistochemistry). (b) diffraction-limited frames taken by a total internal reflection microscope (top), and the reconstructed STORM images of the same frames (bottom). Scale bars, 300 nm.}
    \label{fig:storm}
\end{figure}

High-resolution fluorophore localization can be framed as a linear inverse problem. Let us denote the collected sequence of diffraction-limited frames as $\mathbf{Y}\in \mathds{R}^{M^2 \times T}$, where every column is the $M^2$ vector stacking of the corresponding $M \times M$ frame. Our goal is to reconstruct an image of size $N \times N$, consisting of fluorophore locations on a fine grid ($N>M$). We can model the generation of $\mathbf{Y}$ as:

\begin{equation}
\label{eqn:sc1}
\mathbf{Y=AX},
\end{equation}
where $\mathbf{X}\in \mathds{R}^{N^2 \times T}$ is the sequence of vector-stacked high-resolution frames, and the non-zero entries in each frame (i.e., columns in the matrix) correspond to the locations of activated fluorophores. The matrix $\mathbf{A}\in \mathds{R}^{M^2 \times N^2}$ is the measurement matrix, where each column of $\mathbf{A}$ is defined as the system's PSF shifted by a single pixel on the high-resolution grid.

The simplest way to retrieve $\mathbf{X}$ without leveraging knowledge of its biological structure, is by fitting the observed emission profile, $\mathbf{Y}$, to the PSF of the system, which is typically modelled as a Gaussian function in 2D. This results in localizations with precision greater than the diffraction limit (accurate up to few to tens of nm, versus a diffraction limit of ~200nm), allowing for imaging at a molecular scale within cells. Fig. \ref{fig:storm} illustrates the enhanced resolution of SMLM: STORM reveals the underlying structure of a circular DNA construct, which was completely unseen in its diffraction-limited images.

\begin{figure}
\centering
\includegraphics[width=\textwidth,height=.5\textheight,keepaspectratio]{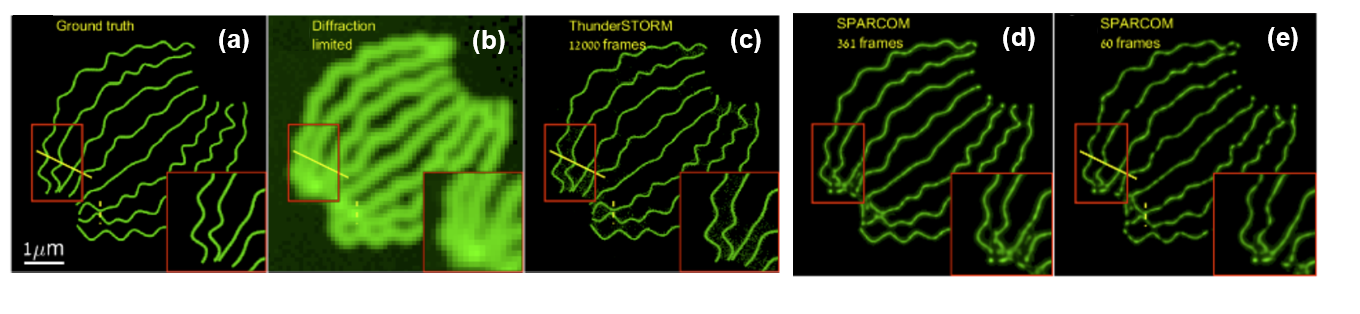}
\caption{\footnotesize Results from \cite{SPARCOM}, showing the simulation and reconstruction of microtubules from a movie \cite{smlm_rev} of 361 high density frames. (a) Simulated ground truth of the image with sub-wavelength features (b) Diffraction-limited image, obtained by summing all the 361 frames in the movie. (c) Single molecule localization reconstruction from a low-density movie of 12,000 frames of the same simulated microtubules and the same number of emitters (the image is constructed using the ThunderSTORM plugin \cite{ThunderSTORM} for ImageJ). SPARCOM recoveries for movies with 361 and 60 high-density frames (simulated microtubules and number of emitters is the same) are given in (d) and (e), respectively.}
\label{fig:sparcom}
\end{figure}

While achieving excellent resolution, standard SMLM methods have one main drawback: they require lengthy imaging times to achieve full coverage of the imaged specimen on the one hand, and minimal overlap between PSFs on the other. Thus, in its classical form this technique has low temporal resolution, preventing its application to fast-changing specimens in live-cell imaging. To circumvent the long acquisition periods required for SMLM methods, a variety of techniques have emerged, which enable the use of a smaller number of frames for reconstructing the 2-D super-resolved image \cite{falcon,CSSTORM,SOFI,SPARCOM,deepSTORM}. These techniques take advantage of prior information regarding either the optical setup, the geometry of the sample, or the statistics of the emitters. One such technique is SPARCOM \cite{SPARCOM_MATH, SPARCOM}, which exploits sparsity in the correlation domain, while assuming that the blinking emitters are uncorrelated over time and space. This allows re-formulation of the localization task as a sparse recovery problem which can be solved using ISTA (see \say{Learned Sparsity-Based Super-Resolution Correlation Microscopy} and Fig. \ref{fig:LSPARCOM} for more details).

SPARCOM yields excellent results when compared to a standard STORM reconstruction (using ThunderSTORM \cite{ThunderSTORM}), as illustrated in Fig. \ref{fig:sparcom}. SPARCOM achieves similar spatial resolution with as few as 361 and even 60 frames, compared with the 12,000 frames needed for ThunderSTORM to produce a reliable recovery, corresponding to a 33- or 200-times faster acquisition rate when using SPARCOM. Thus, SPARCOM improves temporal resolution while retaining the spatial resolution of PALM/STORM. Gaining these benefits comes with tradeoffs: SPARCOM requires prior knowledge of the PSF of the optical setup for the calculation of the measurement matrix, which is not always available, and a careful choice of regularization factor $\lambda$, which is generally done heuristically. 

As shown in the previous section, these shortcomings can be overcome by learning from data using an algorithm unrolling approach. This was done recently by Dardikman-Yoffe et al. \cite{LSPARCOM}, which introduced Learned SPARCOM (LSPARCOM) - a deep network with 10 layers resulting from unfolding SPARCOM, detailed in the box below. 

\vspace{0.5em}
\begin{tcolorbox}[breakable,title={Learned Sparsity-Based Super-Resolution Correlation Microscopy}]
\setstretch{1}

In SPARCOM, we start by observing the temporal covariance matrices of $\mathbf{X}$ and $\mathbf{Y}$, $\mathbf{M}_X$ and $\mathbf{M}_Y$. According to (\ref{eqn:sc1}), we can write the following:

\begin{equation}
\label{eqn:sc3}
\mathbf{M}_Y=\mathbf{AM}_X\mathbf{A}^T.
\end{equation}

We assume that different emitters are uncorrelated over time and space. Thus, $\mathbf{M}_X$ is a diagonal matrix, where each entry on its diagonal, $\mathbf{m}$, represents the variance of the emitter fluctuation on a high-resolution grid. Since non-zero variance can only exist where there is fluctuation in emission, the support of the diagonal corresponds to the emitters' locations on the high resolution grid. Therefore, recovering $\mathbf{m}$ and reshaping it as a matrix yields the desired high-resolution image. For this purpose, let us re-write (\ref{eqn:sc3}) as:

\begin{equation}
\label{eqn:sc4}
\mathbf{M}_Y=\sum_{i=1}^{N^2} \mathbf{A}_i\mathbf{A}^T_i\mathbf{m}_i,
\end{equation}

where $\mathbf{A}_i$ is the $i$-th column in $\mathbf{A}$ and $\mathbf{m}_i$ is the $i$-th entry in $\mathbf{m}$. Following (\ref{eqn:sls}), we can exploit the sparsity of emitters, and compute $\mathbf{m}$ by solving the following sparse recovery problem:

\begin{equation}
\label{eqn:sc5}
\min_{\mathbf{m}\geq0}{\hspace{0.3em}\lambda\|\mathbf{m}\|_1 + \frac{1}{2}\|\mathbf{M}_Y - \sum_{i=1}^{N^2} \mathbf{A}_i\mathbf{A}^T_i\mathbf{m}_i\|^2_2},
\end{equation}

where $\lambda\geq0$ is the regularization parameter. ISTA can be used to solve this optimization problem, as shown in  \ref{fig:LSPARCOM}.

To apply unrolling to SPARCOM, we need to replace the operations performed in a single iteration with neural network layers, and choose the input of the unrolled algorithm. The unrolling process is illustrated in Fig. \ref{fig:LSPARCOM}: to start, $\mathbf{G}$, the $N \times N$ matrix-shaped resized version of the diagonal of $\mathbf{M}_Y$, is taken as input. The matrix-multiplication operations performed in each iteration are replaced with convolutional filters $W^{(k)}_p$, $k$=$0,... ,9$, and the positive soft-thresholding operator is replaced with a differentiable, sigmoid-based approximation of the positive hard-thresholding operator \cite{l0-relu}, denoted as $S^{+}_{\alpha_0, \beta_0}(\cdot)$. The unrolling process results in LSPARCOM - a deep neural network which acts as the operation done by running SPARCOM over multiple iterations. LSPARCOM can be trained on a single sequence of frames taken from one FOV with a known underlying structure, which can be generated using simulations (like the one offered by ThunderSTORM \cite{ThunderSTORM}). The model is then trained on overlapping small patches taken from multiple frames of that sequence.
\includegraphics[width=.9\textwidth,height=.9\textheight,keepaspectratio]{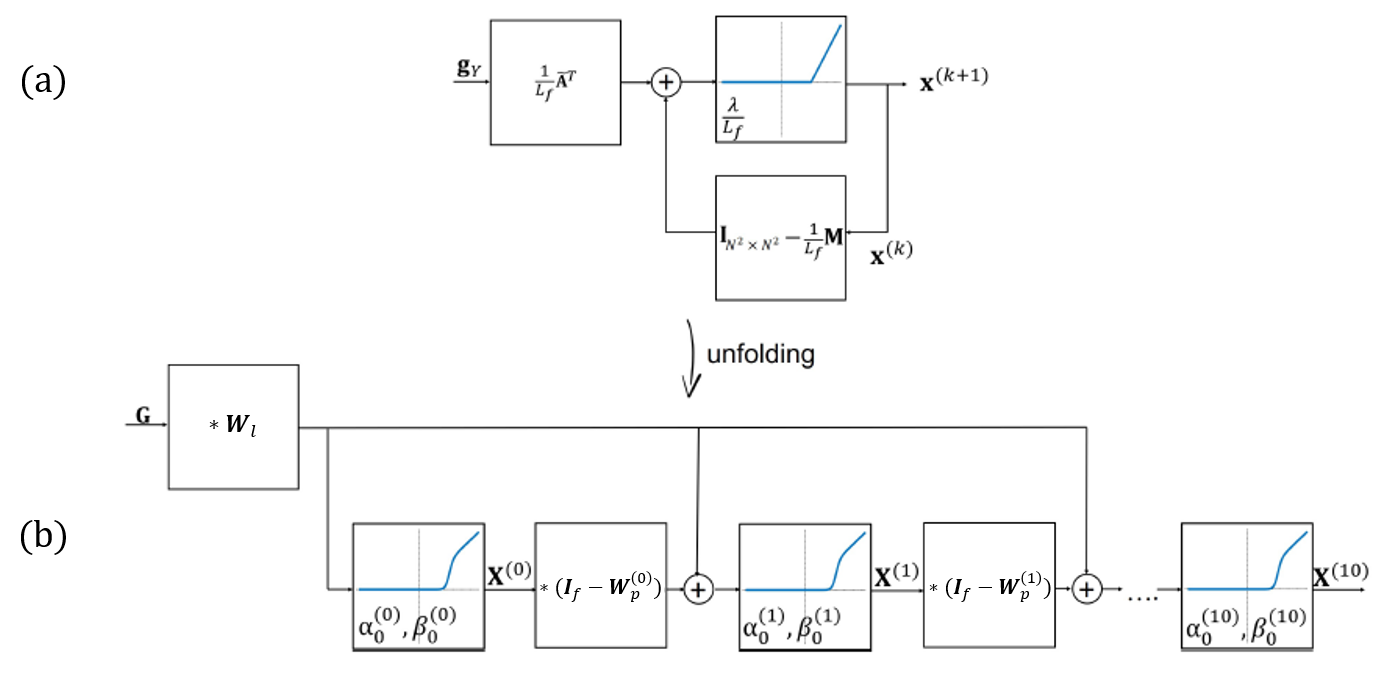}
\captionof{figure}{\footnotesize Unrolling of SPARCOM to LSPARCOM, from \cite{LSPARCOM}. 
(a) block diagram of SPARCOM (via ISTA), recovering the vector-stacked super-resolved image $\mathbf{x}^{(k)}$ (which corresponds to $\mathbf{m}$). The input is $\mathbf{g}_Y$, the diagonal of $\mathbf{M}_Y$. The block with the blue graph is $\mathcal{T}_{\frac{\lambda}{L_f}}$, the positive soft thresholding operator with parameter $\frac{\lambda}{L_f}$, where $L_f$ is the Lipschitz constant of the gradient of (\ref{eqn:sc5}). The other blocks denote matrix multiplication (from the left side), where $\mathbf{\Tilde{A}}=\mathbf{A}^2$ (element-wise power) and $\mathbf{M} = |\mathbf{A}^T\mathbf{A}|^2$ (absolute-value and power operations performed element-wise).
(b) LSPARCOM, recovering the super-resolved image $\mathbf{X}^{(k)}$. The input \textbf{G} is the matrix-shaped resized version of $\mathbf{g}_Y$. The blocks with the blue graph apply the smooth activation function $S^{+}_{\alpha_0, \beta_0}(\cdot)$ with two trainable parameters: $0\geq\alpha_0^{(k)}\geq1$, $\beta_0^{(k)}$, $k$=$0,... ,10$. The other blocks denote convolutional layers, where $I_f$ is a non-trainable identity filter and $W_i$, $W^{(k)}_p$, $k$=$0,... ,9$ are trainable filters.}
\label{fig:LSPARCOM}
\end{tcolorbox}

\begin{figure}[ht]
    \centering
    \includegraphics[width=\textwidth,height=.5\textheight,keepaspectratio]{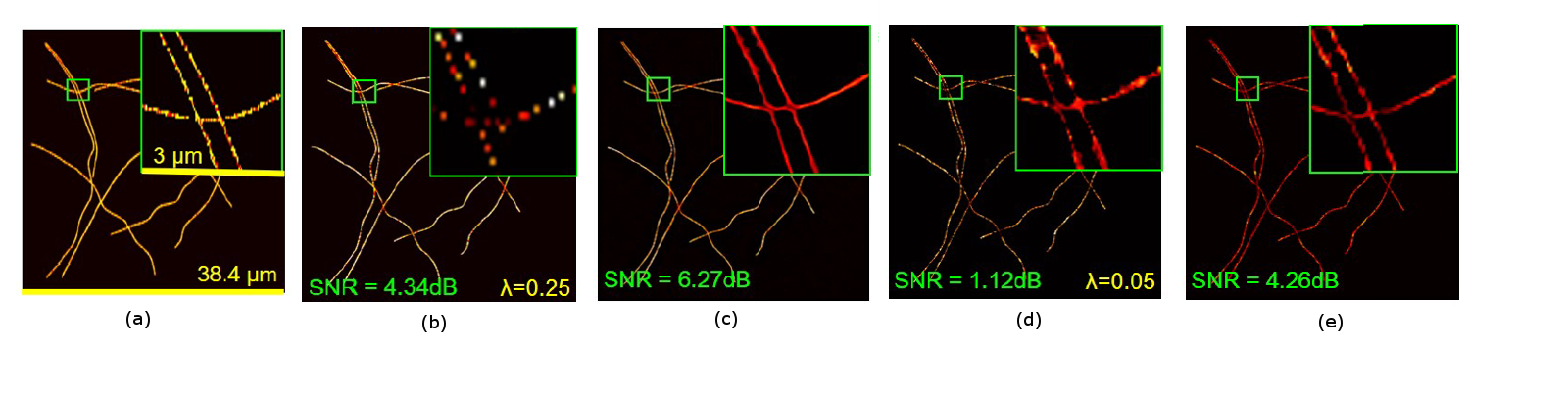}
    \caption{\footnotesize Sample results from \cite{LSPARCOM}, reconstructed from a simulated biological tubulins dataset \cite{smlm_rev}, composed of 350 high-density (b,c) or 25 very high-density frames (d,e). (a): simulated ground truth tubulin structure. (b,d): SPARCOM reconstruction, executed over 100 iterations with $\lambda = 0.25$ for 350 frames (b), and $\lambda = 0.05$ for 25 frames (d). (c,e): LSPARCOM reconstruction, given 350 frames (c) and 25 frames (e) as input.}
    \label{fig:LSPARCOM_results}
\end{figure}

The results shown in Fig. \ref{fig:LSPARCOM_results} illustrate that inference from 10 folds of LSPARCOM is comparable to running SPARCOM for 100 iterations with a carefully-chosen regularization parameter. Both methods succeed in reconstructing the underlying tubulin structure from a sequence of 350 high-density frames. Moreover, if a shorter, denser sequence is constructed by summing groups of 14 frames of the original sequence, the SPARCOM reconstruction's resolution degrades while the LSPARCOM reconstruction remains excellent. Thus, even with 25 extremely dense frames as input, LSPARCOM yields excellent reconstruction of sub-wavelength features, which allows for substantially higher temporal resolution (compared to the hundreds of frames needed for SPARCOM). LSPARCOM is also faster to use, with approximately 5x improvement over SPARCOM in execution time \cite{LSPARCOM}. LSPARCOM enables efficient and accurate imaging well below the diffraction limit, without prior knowledge regarding the PSF or imaging parameters.

Given its enhanced capabilities, LSPARCOM has great potential for localization of biological structures. Meeting the temporal and spatial resolutions' criteria for imaging of dynamic cellular processes at a molecular scale, it might replace its iterative counterpart as a robust, efficient method for live-cell imaging. The success of LSPARCOM further suggests that unrolling may benefit other sparse biological imaging problems, as we discuss in the next section.

\subsection{Optical Microscopy Extensions}

In the previous section, learned unrolling was shown to achieve fast, highly accurate results in SMLM. Here we touch on two applications that may benefit from unrolling: imaging transcriptomics, and synapse detection.

\subsubsection{Imaging Transcriptomics}

\begin{figure}
    \centering
    \includegraphics[scale=0.85]{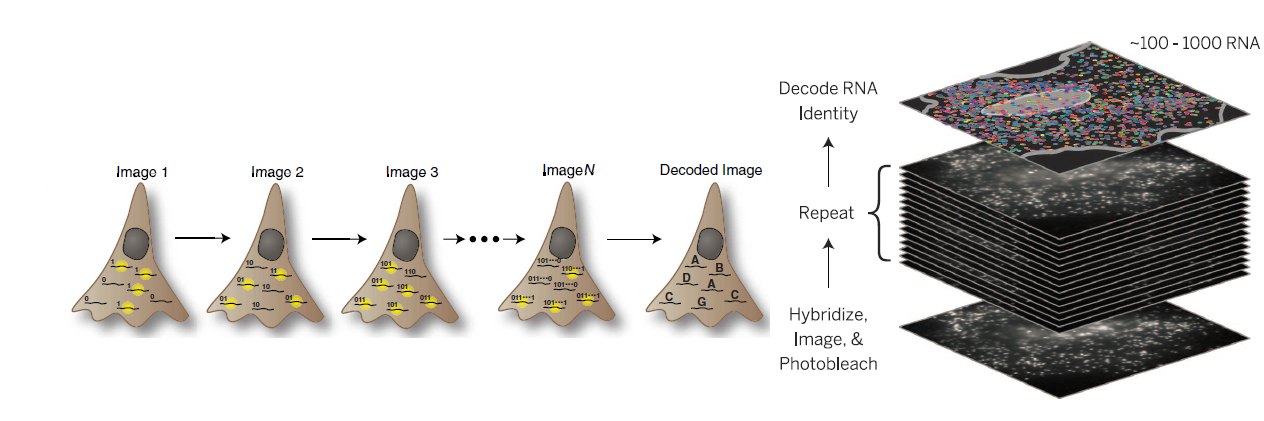}
    \caption{Depiction of MERFISH multiplexed Imaging Transcriptomics, from \cite{MERFISH}. LEFT: Fluorophores are hybridized to mRNA transcripts if the bit of the associated barcode is equal to 1, and mRNA appears as a spot. RIGHT: Acquisition and decoding of MERFISH data. }
    \label{fig:MERFISH}
\end{figure}

Imaging Transcriptomics (IT) is a family of fluorescence microscopy techniques studying the spatial distribution of messenger RNA transcripts in cells. This can allow classification of individual cells by their gene expression in the context of their location in a tissue, yielding insight about the function of the whole system \cite{MERFISH}, or revealing sub-cellular spatial organization of mRNA transcripts. Many IT methods are based on single-molecule Fluorescence \emph{in situ} hybridization (smFISH) \cite{smfish}, in which fluorophore-labeled probes bind to complementary regions of messenger RNA. While smFISH localizes transcripts of one gene at a time, in many experiments, it is desirable to study multiple genes at once, up to tens of thousands. To achieve this goal, combinatorial IT techniques, like MERFISH \cite{MERFISH}, assign a distinct binary barcode with length $F$ to each transcript. The barcodes are chosen to be distinct entries in a \say{codebook}: $F$ rounds of FISH imaging are performed, with transcripts appearing as spots in round $f$ if the $f$-th bit of its barcode is 1, as depicted in Fig. \ref{fig:MERFISH}. By using this technique, up to $2^{F}$ genes may be studied in only $F$ rounds of imaging.

Once these $F$ rounds of imaging are performed, images are processed to produce a set of localizations for mRNA of each gene. The problem of translating images into such a list is a sparse recovery problem: the fluorophores, scattered sparsely across the sample, appear in the images, modulated by the codebook and blurred by the PSF of the microscope. The goal, similarly to SMLM, is to locate these sparsely scattered fluorescent emitters. Currently used processing techniques analyze image data with a heuristic approach in which each location is separately checked for signal. In \cite{JSIT}, we formalized this system analogously to the sparse optimization problem (\ref{eqn:sls}), in a method called the Joint Sparse method for Imaging Transcriptomics (JSIT). For IT data with $F$ rounds of imaging, studying $G$ genes, with $N_{h}$ locations on the high-resolution location grid, and $N_{l}$ pixels in measurement images, we can set up an optimization problem similar to \ref{eqn:sls}. We vectorize and concatenate images into a matrix $\mathbf{Y}\in\mathds{R}^{N_{l}\times F}$. Then, we take $\mathbf{Y}$ to be generated as a product of three matrices, 
\begin{equation}
\label{eqn:JSIT}
    \mathbf{Y} = \mathbf{AXC}.    
\end{equation} 
Here, $\mathbf{A}\in\mathds{R}^{N_{l}\times N_{h}}$, is the same as in (\ref{eqn:ls}), the columns of  $\mathbf{X}\in\mathds{R}^{N_{h}\times G}$ are signal vectors like $\mathbf{x}$ in (\ref{eqn:ls}), with each column representing a specific gene, and $\mathbf{C}\in\mathds{R}^{G\times F}$ is the set of barcodes. Given (\ref{eqn:JSIT}), we recover $\mathbf{X}$ from the measurement $\mathbf{Y}$ and known matrices $\mathbf{A}$ and $\mathbf{C}$, using an optimization-based approach, constrained by assumptions of sparsity: that only one mRNA will be present at each location, and that relatively few mRNA will be present in the FOV. This constrained optimization problem can be solved with an iterative algorithm. In addition to being more interpretable than the currently used heuristic, this method has achieved more accurate mRNA localization, especially in low-magnification imaging.

A natural extension of this formulation is the application of learned unrolling. Based on our experience with other applications, unrolling may improve performance obtaining more accurate genetic expression levels in fewer iterations, while eliminating parameter-tuning requirements and explicit knowledge of the optical PSF $\mathbf{A}$.

\subsubsection{Synapse Detection}
Many other biological settings involve sparse emitters, but have not necessarily been framed as sparse recovery problems to improve performance. For example, sparse biological images are encountered in neuronal synapse detection for characterization of the neurophysiological consequences of genetic and pharmacological perturbation screens. Synapses are localized by identifying positions in which fluorescently tagged pre- and post-synaptic proteins are located in close proximity. These proteins cluster into puncta which appear as point sources, similar to fluorophores in SMLM data. Sparse analysis could identify sub-pixel locations for each punctum, and, from the presence of both pre- and post-synaptic proteins, infer accurate synapse locations. While sparse algorithm unrolling has not yet been applied in this context, model-based learning strategies have already shown good results \cite{dogNet} for this setting.


\section{Unrolling in other imaging modalities}

Sparse emitters arise in other biological imaging modalities beyond epifluorescent microscopy, and the algorithmic unrolling method has achieved fast, highly accurate localization in several such settings. Here we will review three such cases: Ultrasound Localization Microscopy (ULM), Light Field Microscopy (LFM), and cell center localization in non-spatially-sparse histology images.

\subsection{Unrolling in Ultrasound Localization Microscopy}

The attainable resolution of ultrasonography is fundamentally limited by wave diffraction, i.e., the minimum distance between separable scatters is half a wavelength. Due to this limit, conventional ultrasound techniques are bound to a tradeoff between resolution and penetration depth: increases in the transmit frequency shortens the wavelength (thus increasing resolution) but come at the cost of reduced penetration depth, since higher frequency waves suffer from stronger absorption. This tradeoff particularly hinders deep high-resolution microvascular imaging, which is crucial for many diagnostic applications. 

A decade ago, this tradeoff was circumvented by the introduction of Ultrasound Localization Microscopy (ULM) \cite{ULM, ULM2}, which leverages the principles of SMLM and adapts these to ultrasound imaging. In SMLM, stochastic “blinking” of subsets of fluorophores is exploited to provide sparse point sources; in ULM, lipid-shelled gas microbubbles fulfill this role. A sequence of diffraction-limited ultrasonic scans is acquired, each containing just a few active isolated sources. Thus, each received image frame can be written as:

\begin{equation}
\label{eqn:ulm1}
\mathbf{y = Ax + w},
\end{equation}
where $\mathbf{x}$ is a vector that describes the sparse microbubble distribution on a high-resolution image grid, $\mathbf{y}$ is a vectorized image frame from the ultrasound sequence, $\mathbf{A}$ is the measurement matrix (defined by the system's PSF), and $\mathbf{w}$ is a noise vector. As in SMLM, this enables precise localization of their centers on a subdiffraction grid. The accumulation of many such localizations over time yields a super-resolved image. This approach achieves a resolution up to ten times smaller than the wavelength \cite{ufULM},  showing that ultrasonography at sub-diffraction scale is possible. 

Similarly to SMLM, the quality of ULM imaging is dependant on the quantity of localized microbubbles and localization accuracy; thus, it gives rise to a new tradeoff between microbubble density and acquisition time. To achieve the desired signal sparsity for straightforward isolation of the backscattered echoes, ULM is typically performed using a very diluted solution of microbubbles. On regular ultrasound systems, this constraint leads to tediously long acquisition times to cover the full vascular bed. Ultrafast plane-wave ultrasound (uULM) imaging has managed to lower acquisition time \cite{ufULM} by taking many snapshots of individual microbubbles, as they transport through the vasculature, thereby facilitating  high-fidelity reconstruction of the larger vessels. Nevertheless, mapping the full capillary bed still requires microbubbles to pass through each capillary, capping acquisition time benefits to tens of minutes \cite{ULM_TRADEOFF}.

\begin{figure}
    \centering
    \includegraphics[width=.6\textwidth,height=.3\textheight,keepaspectratio]{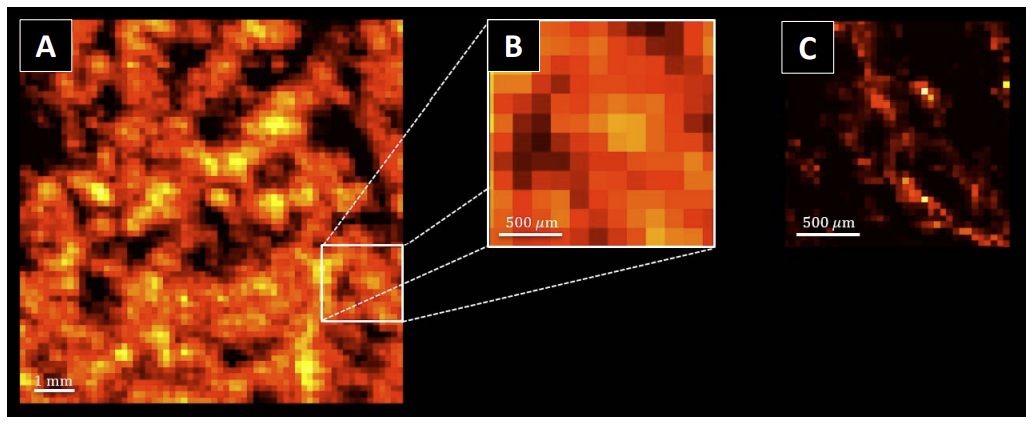}
    \caption{\footnotesize Sample results from \cite{ULM_IMAGE}. Reconstructed highly-dense sequence of 300 frames, clinically acquired \textit{in-vivo} from a human prostate. Maximum intensity projection (MIP) image of the sequence is shown in (A); (B) is a selected area in the image, and (C) is the sparsity-driven super-resolution ultrasound on the same area.}
    \label{fig:ULM_results}
\end{figure}

As with SMLM, uULM can be extended by using the sparsity of the measured signal (whether spatially sparse or in any transform domain \cite{SUSHI}). Sparse recovery again enables improved localization precision and recall for high microbubble concentrations \cite{SPARSE_ULM}. Fig. \ref{fig:ULM_results} illustrates this, showing how the sparse recovery method produces fine visualization of the human prostate from a high-density sequence of \textit{in-vivo} ultrasound scans. However, as in the case of SPARCOM, solving the ULM sparse recovery problem requires iterative algorithms, such as ISTA. Unfortunately, as was previously noted, these algorithms are not computationally efficient, and their effectiveness is strongly dependent on good approximation of the system's PSF and careful tuning of the optimization parameters. With the unrolling approach, these challenges can be met in a similar fashion to SMLM above (see \say{Deep Unrolled ULM} for details).

\vspace{0.5em}
\begin{tcolorbox}[breakable,title={Deep Unrolled ULM}]
\setstretch{1}

Under the assumption of spatial sparsity of microbubbles in the high-resolution grid, $\mathbf{x}$ (as defined in (\ref{eqn:ulm1})) corresponds to the solution of the $l_1$-regularized inverse problem which was previously presented in (\ref{eqn:sls}); thus, it can be computed using ISTA. After estimating $\mathbf{x}$ for each frame, the estimates are summed across all frames to yield the final super-resolution image, which describes the microbubble distribution throughout the entire sequence.

To apply unrolling in this case, LISTA can replace ISTA as in section 2. Van Sloun et al. \cite{DEEP_ULM} have implemented such a model, resulting in a 10-layer feed-forward neural network. Each layer consists of trainable 5x5 convolutional filters $\mathbf{W}_{0k}$ and $\mathbf{W}_k$, along with a trainable shrinkage parameter $\lambda^k$ ($k = 0, ..., 9$). The convolutional filters replace the fully-connected layers which appear in the original unrolled version of ISTA (see Fig. \ref{fig:lista}). Replacing the proximal soft-thresholding operator $T_{\lambda}$ (see (\ref{eqn:th_op})) with a smooth sigmoid-based soft-thresholding operation \cite{l0-relu}, helped avoid vanishing gradients. Similarly to LSPARCOM, this network is trained on simulated ultrasound scans of point sources, with a variety of PSF and noise realizations. Overlapping small patches taken from multiple frames of each simulated scan sequence, and given to the network as training samples. 

\end{tcolorbox}

\begin{figure}
    \centering
    \includegraphics[width=.7\textwidth,height=.25\textheight,keepaspectratio]{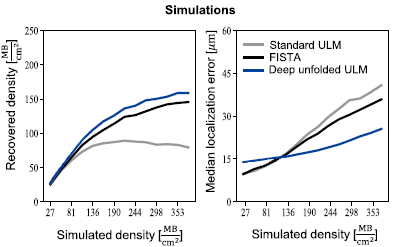}
    \caption{\footnotesize Performance comparison of standard ULM, sparse-recovery (FISTA) and deep unrolled ULM on simulations (taken from \cite{DEEP_ULM}).}
    \label{fig:unfolded_ULM_results}
\end{figure}

Tests on synthetic data show that deep unrolled ULM significantly outperforms standard ULM and sparse recovery through FISTA for high microbubble concentrations (see Fig. \ref{fig:unfolded_ULM_results}), offering better recall (measured by the recovered density) and lower localization error. In addition, when tested on \textit{in-vivo} ultrasound data, Van Sloun et al. \cite{DEEP_ULM} observed that deep unrolled ULM yields super-resolution images with higher fidelity, implying improved robustness and better generalization capabilities. Furthermore, Bar-Shira et al. \cite{breastULM} demonstrated how the use of deep unrolled ULM for \emph{in vivo} human super-resolution imaging allows for better diagnosis of breast pathologies. The unrolled method is also highly efficient, requiring just over 1,000 FLOPS, and containing only 506 parameters (compared to millions of parameters in other deep-learning models).

In sum, deep unrolled ULM can be a method for efficient, robust and parameter-free ultrasonic imaging, with comparable (or superior) resolution to that of other ULM methods (standard and sparsity-based). Given the ability of deep unrolled ULM to perform precise reconstructions for high microbubble concentrations, deep-tissue ultrasound imaging becomes feasible. High microbubble concentrations dramatically shorten required acquisition times, which allows to perform deep high-resolution imaging much faster. Thus, intricate ultrasonography tasks (like microvascular imaging) which have a key role in non-invasive, \textit{in-vivo} diagnosis of many medical conditions such as cancer, arteriosclerosis, stroke and diabetes, become simpler to execute.

\subsection{CISTA}

Another imaging domain dealing with spatially sparse data is Light Field Microscopy (LFM) \cite{LFM}. Obtaining 3D information from a single acquisition is valuable, enabling real-time volumetric neural imaging. LFM enables single-shot 3D imaging by placing a micro-lens array between the microscope objective and the camera sensor. This configuration captures both lateral and angular information from each light ray emitted from the sample, so deconvolution of the system's PSF produces 3D emitter locations. Since each spatial location is imaged in multiple pixels on the detector, LFM faces a tradeoff between depth and lateral resolution. However, if the sample is composed of spatially sparse emitters, localization on a high-resolution, 3D grid can be performed, as in SMLM and ULM. 

\begin{figure}
    \centering
    \includegraphics[scale=0.75]{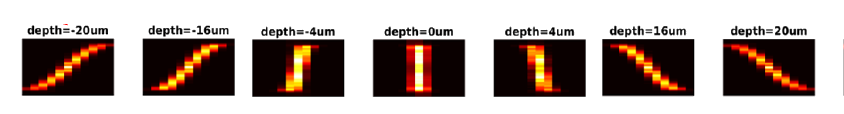}
    \caption{Epipolar Plane Images (EPI) derived from LFM images of emitters at different depths (from \cite{LFM_CS} under Creative Commons License 4.0). By matching an observed EPI, the depth of sources may be determined.}
    \label{fig:EPI}
\end{figure}

In \cite{CISTA}, the problem of localizing neurons in 3D space with LFM images is presented: in LFM, neurons are small enough to be considered point sources, and are distributed in a spatially sparse manner. To localize neurons, measured images are converted to a structure called an Epipolar Plane Image (EPI); the system PSF in this domain varies strongly with depth, as shown in Fig. \ref{fig:EPI}. By performing sparse optimization, the authors are able to achieve fast, accurate neuron localization. 

By framing 3D neuron localization as a sparse optimization problem, the problem is opened to unrolling. In \cite{CISTA}, Song et al. first use a convolutional variant of ISTA (CISTA) to solve the localization problem, then create an unrolled network based on that algorithm, called CISTA-net. The unrolled network recovers neuron location with higher accuracy in all dimensions and performs the recovery task more than 10,000 times faster than ISTA. This increase in speed expands the applicability of LFM: it could enable, for instance, live, 3D imaging of whole nervous systems in small model organisms like \emph{C. elegans}, or of activity of large volumes of the mammalian cortex.

\subsection{Non-Spatially Sparse Imaging}

The most obvious way of thinking about sparse recovery in biological imaging is in the domain of spatially sparse sources, but there are other methods which leverage sparse coding in other aspects, and unrolling can achieve accurate results in these situations as well.

One example is cell center localization in histology slides. While cell centers are scattered sparsely in a FOV, cell shapes are irregular, so there is not a single \say{impulse response} transforming cell center locations into images of cells, as in the sparse recovery form of (\ref{eqn:sls}). In \cite{ecnncs}, a traditional CNN is combined with a LISTA-like network to localize cell centers. In this framework, the locations of the centers of cells in a 2-dimensional FOV with dimension $h \times w$ are represented by a binary matrix $\mathbf{X}\in\mathds{R}^{h\times w}$. The matrix $\mathbf{X}$ is Radon transformed to represent the cell centers in polar coordinates, $\mathbf{X}_{p} = \mathcal{R}f(X)$. A measurement matrix $\mathbf{A}$ is generated as a random Gaussian projection matrix, and the product $\mathbf{AX_{p}} = \mathbf{Y}$ is formed. Xue et al. found that although $\mathbf{Y}$ cannot be measured directly, a CNN may be trained to infer $\mathbf{Y}$ from an image of the FOV. A two-stage neural network is designed: the first stage a traditional CNN, transforming the images into an estimate $\mathbf{\hat{Y}}$ of the matrix $\mathbf{Y}$, the second stage a LISTA-like network used to obtain an estimate $\mathbf{\hat{X}_{p}}$ of the sparse matrix $\mathbf{X_{p}}$, which, after inverse Radon transformation, gives the cell center locations $\mathbf{X}$. The network, called End-to-end Convolutional Neural Network and Compressed Sensing (ECNNCS), is trained by penalizing the differences between both $\mathbf{\hat{y}}$ and $\mathbf{y}$ and $\mathbf{\hat{x}}$ and $\mathbf{x}$. The ECNNCS model achieved better localization accuracy than the state-of-the-art algorithms used as comparison \cite{ecnncs}, showing that unrolling can improve performance outside problems of strict spatial sparsity. 

Another technique in which sparse recovery may be applied to biological imaging data which is not spatially sparse is Compressed \emph{in situ} Imaging (CISI) \cite{cisi}, and we propose that algorithm unrolling will improve performance in this method as well. Like Imaging Transcriptomics (IT), CISI is a microscopy technique which evaluates expression levels of genes at single-cell resolution. Different from IT, in CISI data are not spatially sparse. Instead, CISI takes advantage of genetic co-expression patterns to infer single cells' transcriptomes from a few measurements of multiple genes at once (\say{composite measurements}). In CISI, single-cell transcriptomes are conceptualized as linear combinations of \say{modules}, (sparse) linear combinations of co-expressed genes. The sparse recovery problem is to infer from composite measurements of genes the sparse set of active co-expression modules. Currently, modules are defined before the experiment, but with algorithmic unrolling, optimal co-expression modules could be learned, enabling improved transcriptome inference.


\section{Conclusion}

New biological imaging techniques are constantly being developed, and with them, computational pipelines to identify and characterize the imaged biological structures. We have described a few of these techniques and their accompanying pipelines. In many cases, these techniques consist of heuristic strategies which have limited accuracy and are difficult to interpret. As computational power continues to increase and the methods become more developed, more powerful, interpretable processing techniques have been created by incorporating biological and physical assumptions into constrained optimization problems, solved with iterative methods. These in turn require parameter tuning and explicit knowledge of the experimental setup. A natural next step in pipeline development is model-based learning methods, including algorithmic unrolling. We have shown how, in many imaging modalities requiring source localization, unrolling achieves fast, accurate results with robust models, and proposed that unrolling be extended widely to other similar problems, including to methods involving biological structure other than sparsity. We hope this work will inspire methods extending unrolling to further biological imaging modalities and experimental settings.


\bibliographystyle{IEEEbib}
\setstretch{0.8}
\bibliography{myRef}

\begin{thebibliography}{10}

\bibitem{storm_09}
X.~Zhuang,
\newblock ``Nano-imaging with {STORM},''
\newblock {\em Nature photonics}, vol. 3, no. 7, pp. 365--367, 2009.

\bibitem{smfish}
A.M. Femino, F.S. Fay, K.~Fogarty, and R.H. Singer,
\newblock ``Visualization of single {RNA} transcripts in situ,''
\newblock {\em Science}, vol. 280, pp. 585--590, 1998.

\bibitem{dogNet}
V.~Kulikov, S.-M. Guo, M.~Stone, A.~Goodman, A.~Carpenter, M.~Bathe, and
  V.~Lempitsky,
\newblock ``Dognet: A deep architecture for synapse detection in multiplexed
  fluorescence images,''
\newblock {\em PLOS Computational Biology}, vol. 15, 2019.

\bibitem{ecnncs}
Y.~Xue, G.~Bigras, J.~Hugh, and N.~Ray,
\newblock ``Training convolutional neural networks and compressed sensing
  end-to-end for microscopy cell detection,''
\newblock {\em IEEE transactions on medical imaging}, vol. 38, no. 11, pp.
  2632--2641, 2019.

\bibitem{MERFISH}
K.H. Chen, A.N. Boettiger, J.R. Moffitt, S.~Wang, and X.~Zhuang,
\newblock ``Spatially resolved, highly multiplexed {RNA} profiling in single
  cells,''
\newblock {\em Science}, vol. 348, pp. 412, 2015.

\bibitem{LSPARCOM}
G.~Dardikman-Yoffe and Y.~C. Eldar,
\newblock ``Learned {SPARCOM}: Unfolded deep super-resolution microscopy,''
\newblock {\em Optics Express}, vol. 28-19, pp. 27736 -- 27763, September 2020.

\bibitem{smlm_rev}
D.~Sage, H.~Kirshner, T.~Pengo, N.~Stuurman, J.~Min, S.~Manley, and M.~Unser,
\newblock ``Quantitative evaluation of software packages for single-molecule
  localization microscopy,''
\newblock {\em Nature methods}, vol. 12, no. 8, pp. 717--724, 2015.

\bibitem{JSIT}
J.~Bryan, B.~Cleary, S.~Farhi, and Y.C. Eldar,
\newblock ``Sparse recovery of imaging transcriptomics data,''
\newblock {\em Proceedings of the International Symposium on Biomedical
  Imaging}, 2021.

\bibitem{ULM}
I.~Grundberg, S.~Kiflemariam, M.~Mignardi, K.~Imgenberg, K.~Edlund, P.~Micke,
  M.~Sundström, T.~Sjöblom, J.~Botling, and M.~Nilsson,
\newblock ``Super-resolution ultrasound imaging,''
\newblock {\em Ultrasound in medicine and biology}, vol. 46, pp. 865–891,
  2020.

\bibitem{CISTA}
P.~Song, H.~V. Jadan, C.~L. Howe, P.~Quicke, A.~J. Foust, and P.~L. Dragotti,
\newblock ``Model-inspired deep learning for light-field microscopy with
  application to neuron localization,''
\newblock {\em arXiv preprint arXiv:2103.06164}, 2021.

\bibitem{LISTA}
K.~Gregor and Y.~LeCun,
\newblock ``Learning fast approximations of sparse coding,''
\newblock {\em Proc. of 27th Int. Conf. on machine learning}, pp. 399 -- 406,
  2010.

\bibitem{Unrolling_rev}
V.~Monga, Y.~Li, and Y.~C. Eldar,
\newblock ``Algorithm unrolling: Interpretable, efficient deep learning for
  signal and image processing,''
\newblock {\em IEEE Signal Processing Magazine}, pp. 17--43, 2021.

\bibitem{LASSO}
R.~Tibshirani,
\newblock ``Regression shrinkage and selection via the lasso,''
\newblock {\em Journal of the Royal Statistical Society. Series B
  (Methodological)}, vol. 58-1, pp. 267 -- 288, 1996.

\bibitem{ADMM}
S.~Boyd, N.~Parikh, E.~Chu, B.~Peleato, and J.~Eckstein,
\newblock ``Distributed optimization and statistical learning via the
  alternating direction method of multipliers,''
\newblock {\em Foundations and Trends in Machine Learning}, vol. 3, pp. 1--122,
  2011.

\bibitem{ISTA}
I.~Daubechies, M.~Defrise, and C.De Mol,
\newblock ``An iterative thresholding algorithm for linear inverse problems
  with sparsity constraint,''
\newblock {\em Communications on on Pure and Applied Mathematics}, vol. 57-11,
  pp. 1413--1457, November 2004.

\bibitem{HQS-net}
K.~Zhang, L.~V. Gool, and R.~Timofte,
\newblock ``Deep unfolding network for image super-resolution,''
\newblock in {\em Proceedings of the IEEE/CVF Conference on Computer Vision and
  Pattern Recognition}, 2020, pp. 3217--3226.

\bibitem{deepSTORM}
E.~Nehme, L.~E. Weiss, T.~Michaeli, , and Y.~Shechtman,
\newblock ``Deep-{STORM}: super-resolution single-molecule microscopy by deep
  learning,''
\newblock {\em Optica}, vol. 5, pp. 458--464, 2018.

\bibitem{ADMM-net}
Y.~Yang, J.~Sun, H.~Li, and Z.~Xu,
\newblock ``Deep {ADMM-Net} for compressive sensing {MRI},''
\newblock in {\em Proceedings of the 30th International Conference on Neural
  Information Processing Systems}, Red Hook, NY, USA, 2016, NIPS'16, p.
  10–18, Curran Associates Inc.

\bibitem{rPCA}
O.~Solomon, R.~Cohen, Y.~Zhang, Y.~Yang, Q.~He, J.~Luo, R.~JG van Sloun, and
  Y.~C. Eldar,
\newblock ``Deep unfolded robust {PCA} with application to clutter suppression
  in ultrasound,''
\newblock {\em IEEE transactions on medical imaging}, vol. 39, no. 4, pp.
  1051--1063, 2019.

\bibitem{iminpaint}
Y.~Li, M.~Tofighi, J.~Geng, V.~Monga, and Y.~C. Eldar,
\newblock ``Efficient and interpretable deep blind image deblurring via
  algorithm unrolling,''
\newblock {\em IEEE Transactions on Computational Imaging}, vol. 6, pp.
  666--681, 2020.

\bibitem{sub_diff}
E.~Betzig,
\newblock ``Proposed method for molecular optical imaging,''
\newblock {\em Optics Letters}, vol. 20, pp. 237--239, 1995.

\bibitem{SMLM1}
E.~Betzig, G.~H. Patterson, R.~Sougratand O.~W. Lindwasser, S.~Olenych, J.~S.
  Bonifacino, M.~W. Davidson, J.~Lippincott-Schwartz, and H.~F. Hess,
\newblock ``Imaging intracellular fluorescent proteins at nanometer
  resolution,''
\newblock {\em Science}, vol. 313, pp. 1642--1645, September 2006.

\bibitem{SMLM2}
M.~J. Rust, M.~Bates, and X.~Zhuang,
\newblock ``Sub-diffraction-limit imaging by stochastic optical reconstruction
  microscopy ({STORM}),''
\newblock {\em Nature Methods}, vol. 3, pp. 793--796, August 2006.

\bibitem{SPARCOM}
O.~Solomon, M.~Mutzafi, M.~Segev, and Y.~C. Eldar,
\newblock ``Sparsity-based super-resolution microscopy from correlation
  information,''
\newblock {\em Optics Express}, vol. 26-14, pp. 18238--18269, June 2018.

\bibitem{ThunderSTORM}
M.~Ovesný, P.~Křížek, J.~Borkovec, Z.~Svindrych, and G.~M. Hagen,
\newblock ``{ThunderSTORM}: a comprehensive imagej plug-in for {PALM} and
  {STORM} data analysis and super-resolution imaging,''
\newblock {\em Bioinformatics}, vol. 30, no. 16, pp. 2389--2390, 2014.

\bibitem{falcon}
J.~Min, C.~Vonesch, H.~Kirshner, L.~Carlini, N.~Olivier, S.~Holden, S.~Manley,
  J.~C. Ye, and M.~Unser,
\newblock ``{FALCON:} fast and unbiased reconstruction of high-density
  super-resolution microscopy data,''
\newblock {\em Scientific Reports}, vol. 4, pp. 1--9, 2014.

\bibitem{CSSTORM}
L.~Zhu, W.~Zhang, D.~Elnatan, and B.~Huang,
\newblock ``Faster {STORM} using compressed sensing,''
\newblock {\em Nature Methods}, vol. 9, pp. 721--726, 2012.

\bibitem{SOFI}
T.~Dertinger, R.~Colyer, G.~Iyer, S.~Weiss, and J.~Enderlein,
\newblock ``Fast, background-free, {3D} super-resolution optical fluctuation
  imaging ({SOFI}),''
\newblock {\em Proceedings of the National Academy of Sciences of the United
  States of America}, vol. 106(52), pp. 22287--22292, 2009.

\bibitem{SPARCOM_MATH}
O.~Solomon, Y.~C. Eldar, M.~Mutzafi, and M.~Segev,
\newblock ``{SPARCOM}: Sparsity based super-resolution correlation
  microscopy,''
\newblock {\em SIAM J. on Img. Sci.}, vol. 12-1, pp. 392--419, Feb. 2019.

\bibitem{l0-relu}
X.P. Zhang,
\newblock ``Thresholding neural network for adaptive noise reduction,''
\newblock {\em IEEE transactions on neural networks}, vol. 10, pp. 567--584,
  2001.

\bibitem{ULM2}
Y.~Desailly, O.~Couture, M.~Fink, and M.~Tanter,
\newblock ``Sono-activated ultrasound localization microscopy,''
\newblock {\em Applied Physics Letters}, vol. 103, pp. 174107, 2013.

\bibitem{ufULM}
C.~Errico, J.~Pierre, S.~Pezet, Y.~Desailly, Z.~Lenkei, O.~Couture, and
  M.~Tanter,
\newblock ``Ultrafast ultrasound localization microscopy for deep
  super-resolution vascular imaging,''
\newblock {\em Nature}, vol. 527, pp. 499–502, 2015.

\bibitem{ULM_TRADEOFF}
V.~Hingot, C.~Errico, B.~Heiles, L.~Rahal, M.~Tanter, and O.~Couture,
\newblock ``Microvascular flow dictates the compromise between spatial
  resolution and acquisition time in ultrasound localization microscopy,''
\newblock {\em Scientific Reports}, vol. 9, 2019.

\bibitem{ULM_IMAGE}
O.~Solomon, R.~van Sloun, H.~Wijkstra, M.~Mischi, and Y.~C. Eldar,
\newblock ``Sparsity-driven super-resolution in clinical contrast-enhanced
  ultrasound,''
\newblock {\em IEEE International Ultrasonics Symposium}, pp. 1--4, 2017.

\bibitem{SUSHI}
A.~Bar-Zion, O.~Solomon, C.~Tremblay-Darveau, D.~Adam, and Y.~C. Eldar,
\newblock ``{SUSHI}: Sparsity-based ultrasound super-resolution hemodynamic
  imaging,''
\newblock {\em IEEE Transactions on Ultrasonics, Ferroelectrics, and Frequency
  Control}, vol. 65, pp. 2365–2380, 2018.

\bibitem{SPARSE_ULM}
O.~Solomon, R.~van Sloun, H.~Wijkstra, M.~Mischi, and Y.~C. Eldar,
\newblock ``Exploiting flow dynamics for superresolution in contrast-enhanced
  ultrasound,''
\newblock {\em IEEE Transactions on Ultrasonics, Ferroelectrics, and Frequency
  Control}, vol. 66, pp. 1573--1586, 2019.

\bibitem{DEEP_ULM}
R.~van Sloun, R.~Cohen, and Y.~C. Eldar,
\newblock ``Deep learning in ultrasound imaging,''
\newblock {\em Proceedings of the IEEE}, vol. 108, pp. 11--29, 2020.

\bibitem{breastULM}
O.~Bar-Shira, A.~Grubstein, Y.~Rapson, D.~Suhami, E.~Atar, K.~Peri-Hanania,
  R.~Rosen, and Y.~C. Eldar,
\newblock ``Learned super resolution ultrasound for improved breast lesion
  characterization,''
\newblock {\em International Conference on Medical Image Computing and Computer
  Assisted Intervention}, pp. 1--10, 2021.

\bibitem{LFM}
M.~Levoy, R.~Ng, A.~Adams, M.~Footer, and M.~Horowitz,
\newblock ``Light field microscopy,''
\newblock in {\em ACM SIGGRAPH 2006 Papers}, pp. 924--934. 2006.

\bibitem{LFM_CS}
P.~Song, H.~V. Jadan, C.~L. Howe, P.~Quicke, A.~J. Foust, and P.~L. Dragotti,
\newblock ``{3D} localization for light-field microscopy via convolutional
  sparse coding on epipolar images,''
\newblock {\em IEEE Transactions on Computational Imaging}, vol. 6, pp.
  1017--1032, 2020.

\bibitem{cisi}
B.~Cleary, B.~Simonton, J.~Bezney, E.~Murray, S.~Alam, A.~Sinha, E.~Habibi,
  J.~Marshall, E.~S. Lander, F.~Chen, and A.~Regev,
\newblock ``Compressed sensing for imaging transcriptomics,''
\newblock {\em bioRxiv}, 2020.

\end{thebibliography}

\end{document}